\documentclass{article}
\usepackage{natbib}
\usepackage[final]{neurips_2024}
\usepackage[utf8]{inputenc} 
\usepackage[T1]{fontenc}    
\usepackage{hyperref}       
\usepackage{url}            
\usepackage{booktabs}       
\usepackage{amsfonts}       
\usepackage{nicefrac}       
\usepackage{microtype}      
\usepackage{xcolor}         
\usepackage[english]{babel}
\usepackage[utf8]{inputenc}
\usepackage{graphicx}
\usepackage{epsfig}
\usepackage{amsmath}
\usepackage{float}
\usepackage{algorithm}
\usepackage{algpseudocode}

\title{From Correlation to Causation: Understanding Climate Change through Causal Analysis and LLM Interpretations}

\author{
  Shan Shan\thanks{NeurIPS 2024 Workshop on Causality and Large Models(CaLM)} \\Zhejiang University\\
  \texttt{shshan@zju.edu.cn} \\
}

\begin{document}

\maketitle

\begin{abstract}

This research presents a three-step causal inference framework that integrates correlation analysis, machine learning-based causality discovery, and LLM-driven interpretations to identify socioeconomic factors influencing carbon emissions and contributing to climate change. The approach begins with identifying correlations, progresses to causal analysis, and enhances decision making through LLM-generated inquiries about the context of climate change. The proposed framework offers adaptable solutions that support data-driven policy-making and strategic decision-making in climate-related contexts, uncovering causal relationships within the climate change domain.

\end{abstract}

\section{Introduction}

Why do we seek to precisely understand causality? In the realm of large-scale, high-dimensional datasets, is mere knowledge of correlations sufficient? Does our pursuit of causality stem from mere curiosity, or does it offer substantial practical benefits? Could our perceived understanding of causality, much like Plato's allegory of the shadows in the cave, actually obscure the true nature of reality? As Ludwig Wittgenstein notes in his \textit{Philosophical Investigations}, "The precise and explicit rules governing the logical structure of propositions often serve as a concealed backdrop within our medium of understanding." He further discusses the "crystalline purity of logic," highlighting its indispensable role not merely as an outcome of inquiry, but as a foundational necessity \cite[4f,c]{wittgenstein1967philosophical};\cite[49-50]{sluga1996cambridge}. Expanding on this framework of logic structure in understanding, Judea Pearl fosters the concept of "understanding" as a means to the sensation of control, specifically through causal inference which he associates with decision-making in intelligent systems \cite{pearl2014probabilistic}. He posits that a robust understanding of causality is crucial for effective decision-making in intelligent systems, emphasizing that such understanding goes beyond mere data correlation and involves the ability to manipulate and control outcomes\footnote{The Science of Cause and Effect: From Deep Learning to Deep Understanding, \url{https://simons.berkeley.edu/}}. 

These perspectives are particularly relevant to the development of Large Language Models (LLMs), where design elements like prompts are tailored to reflect human-machine interaction. The architecture of LLMs not only displays the capacity of machines to emulate complex human logical processes but also enables further exploration of causal relationships \citep{jin2024largelang,kiciman2024causal,ceraolo2024}. LLMs' prompted and related design highlights the potential of machines to emulate human logical processes and probe into causal relationships of deep understanding.  

However, recent studies have indicated that LLMs are "weak causal parrots", merely reciting the causal knowledge from the training data\citep{zečević2023causalparrotslargelanguage}, parroting unintentional remarks. The primary challenges in causality research arise from the lack of a clear definition of causal statements and the absence of adequate mathematical tools to address these complex questions \citep{pearl2000models}. This research confronts these challenges by comparing and integrating three approaches: correlation analysis, machine learning-based causality discovery, and LLM-inquiry-driven interpretations, within a human-machine interaction framework focused on the social context of climate change. The progression emphasizes not only the identification of correlations and causal factors but also leverages earlier stages and prior knowledge as benchmarks for advanced logical inquiries and interpretations.

The study aims to delineate the effectiveness of above three approaches within the social science context of climate change, striving to deepen our understanding of causality through a comparative framework. By exploring and progressing through three stages, this research seeks to uncover insights that could lead to more informed decisions and strategies for a deeper understanding of social factors that influence climate change. In summary, the study's contributions are as follows:

\begin{itemize}

\item\textbf {Comparative Methodological Framework:} The research uses a three-step comparative framework that combines correlation analysis, machine learning-based causality discovery, and LLM-prompt-driven inquiries (See Appendix Figure \ref{fig:FigA1}) . This approach improves the study of causality and provides a structured way to assess different methods, offering a framework that could be replicated in future research in a similar climate change context.

\item\textbf {Application to Climate Change Social Contexts:} The study focuses on the social science aspects of climate change by exploring correlations and causations of social influences on climate change. The research continues to explore how LLMs can understand and help address these critical social issues, providing valuable insights that could shape environmental policies and strategies, thus improving decision making.

\end{itemize}

\section{Preliminaries: Causal Relations}

It is already known that correlation does not imply causation\cite{ksir2016correlation,rohrer2018thinking}. However, causation is a subset of correlation because a causal relationship inherently implies correlation (but with a cause-and-effect dynamic). Literarily and technically, when exploring causation, correlation represents a closer relationship between two variables than non-relation. Causation cannot exist without correlation, even though correlation alone is not sufficient to establish causation. Following this logical sequence, this research begins by understanding the desired outcome and then determining the necessary steps to achieve it. The study assumes that correlation analysis does not need to be excluded; instead, it could serve as a foundation to narrow down the selection of relationships. 

In complex real-world scenarios, identifying the associations between events and variables helps predict outcomes, make informed decisions, and understand the underlying mechanisms of systems \citep{pearl2000models}. Causal discovery involves identifying the dependent variables of an event of interest and understanding the physical influence relationships between events or variables. Causal structures imply both statistical (conditional) independence and independences to other (non-statistical) information measures \citep{peters2017elements}, which is a common task in causal inference. In the domain of causal discovery, machine learning algorithms can primarily be divided into two main categories: 

\begin{itemize}

\item\textbf{Constraint-based Algorithms}: These algorithms rely on tests of statistical independence within the data to uncover potential causal relationships between variables. They attempt to construct causal graphs by analyzing the conditional independencies among variables. A classic method is the PC algorithm (Peter-Clark Algorithm), which iteratively examines and eliminates edges that do not satisfy conditional independencies, thereby inferring the causal structure between variables \citet{spirtes2001causation}. 

\item\textbf{ Score-based Algorithms:} This type of algorithm identifies the best causal graph by assigning scores to different causal models. The process typically involves enumerating and scoring various possible causal graphs, selecting the model with the highest score. Scoring criteria may be based on how well the data fits the model, such as the Bayesian Information Criterion (BIC) or the Minimum Description Length (MDL). The core idea behind these algorithms is that the causal model which best explains the observed data is considered optimal \citep{liu2012empirical,nogueira2022methods}.
\end{itemize}

Constraint-based algorithms are generally more efficient as they rely on statistical tests to quickly narrow down the search space, but they may be sensitive to noise in the data and sample size. On the other hand, score-based algorithms, though theoretically capable of exploring a broader model space to find the optimal model, can be computationally expensive in practice due to the need to evaluate a large number of models. Overall, the search over causal graphs between variables is challenged by two distinct factors: the sheer volume of causal graphs, which increases super-exponentially with the addition of nodes, and the constraint of maintaining acyclicity\citep{cheng2024data}.

To address this limitation, Rolland et al.  \citep{rolland2022score} design a novel order-based methods to recover causal graphs from the score of the data distribution in non-linear additive noise models and propose a new efficient method for approximating the score’s Jacobian, enabling to recover the causal graph. Specifically, they first sequentially identify leaves of the causal graph by analyzing its entailed observational score, and then remove the identified leaf variables. As a result, one can obtain a complete topological order with a time complexity linear in the number of nodes. Since the node in the ordering can be a parent only of the nodes appearing after it in the same ordering, once a topological order is fixed, the acyclicity constraint is naturally enforced, making the pruning step easier to solve.

\section{Correlation Analysis: Narrowing the Variable Pool}
\begin{figure}[H]
    \centering
    \includegraphics[width=0.8\linewidth]{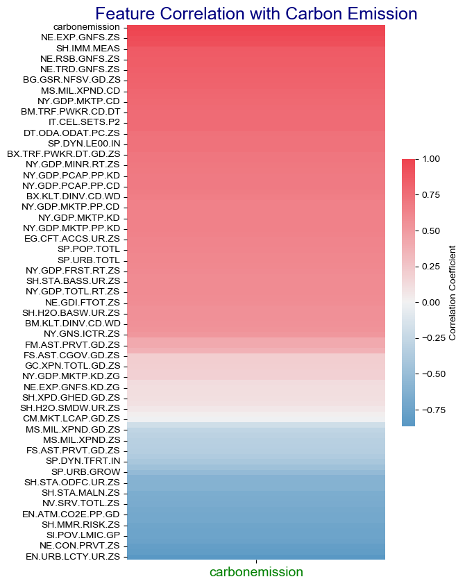}
    \caption{\textbf{Ordered Feature Correlation with Carbon Emissions.} This figure shows the correlation between various social and economic factors and carbon emissions per capita, highlighting key influences such as energy use, GDP, urban population, and access to clean technologies.}
    \label{fig:figA3}
\end{figure}

The correlation step involves correlation analysis, using a heatmap and the Anderson-Darling k-sample test (anderson\_ksamp) with a threshold value of 0.1 (i.e. 10\%) to identify the influence of each variable on the target variable, carbon emissions. Variables with matching distributions between the training and testing data are retained. The most relevant variables are then carried forward to the causation analysis.
\begin{itemize}

\item\textbf{Correlation Matrix Calculation}: First, a correlation matrix is calculated for the dataset, quantifying the linear relationships between each pair of variables. The matrix visually distinguishes positive from negative correlations, with coefficients ranging from "-1" strong negative correlation to "+1" strong positive correlation. The full lists of correlation map are presented in Appendix Figure \ref{fig:figA2}.

\item\textbf{Heatmap Visualization}: Second, the heatmap incorporates hierarchical clustering, which groups variables with similar correlation patterns together. This step enhances the interpretability of the heatmap by organizing it into blocks of highly correlated variables, making it easier to spot clusters of factors that behave similarly. 

The resulting correlation map provides a visual summary of how different factors related to carbon emissions interact with each other. It reveals key drivers of carbon emissions and potential areas for further investigation or intervention, identifying the relationships within complex datasets, and facilitating deeper insights into the underlying dynamics of carbon emissions. Appendix Figure \ref{fig:figA3} shows the sorted correlations of features with the target variable, carbon emissions.
\end{itemize}

\section{Causal Effects Estimation}

The method employed in this research is adapted from existing approaches to causal modeling, specifically following the framework outlined by Rolland et al. (2022) \citep{rolland2022score}. In this approach, each variable is modeled as a function of its direct causal parents in the causal graph, along with an additive noise term. The data distribution is defined by these causal relationships, and score functions are used to identify leaf nodes within the graph. Leaf nodes are detected based on the variance of partial derivatives of the score function, which helps distinguish parent-child relationships among variables. The nodes in the graph are arranged in order by finding and removing leaf nodes one by one. To do this, the experiment uses the Stein gradient estimator with ridge RBF kernel regression to estimate the score function. 

\subsection{Causal Graph Construction and Score Matching}

Take the dataset \(\{V^{2000}, Y^{2005}\}\) as an illustration example, which treats as 16 study variables \(X_{1 \ldots 16} = \{V^{2000}, Y^{2005}\}\). The study assumes the data is generated using the following model:
\[
X_i = f_i(\text{pa}_i(X)) + \epsilon_i, \quad i \in \{1, 2, \ldots, 16\},
\]
where \(\text{pa}_i(X)\) selects the coordinates of \(X\) that are direct causal variables of node \(i\) in the causal graph, and \(\epsilon_i\) is an additive noise term, which might include measurement errors. The associated probability distributions are given by:
\[
p(x) = \prod_{i=1}^d p(x_i \mid \text{pa}_i(x))
\]
\[
\log p(x) = \sum_{i=1}^d \log p(x_i \mid \text{pa}_i(x))
\]
The score function is defined as:
\[
s(x) \equiv \nabla \log p(x)
\]
The necessary and sufficient conditions for the \(j\)-th variable to be a leaf node are given by:
\[
\forall x, \left(\frac{\partial s_j(x)}{\partial x_j}\right) = c, \quad \text{where } c \text{ is a constant value independent of } x,
\]
\[
\text{Var}_X \left(\frac{\partial s_j(X)}{\partial x_j}\right) = 0
\]
If the \(j\)-th variable is a leaf node, and the \(i\)-th variable is its parent node, then:
\[
\text{Var}_X \left(\frac{\partial s_j(X)}{\partial x_i}\right) \neq 0
\]
Based on this finding provided by Rolland et al., 2022, the experiment achieves topological ordering by sequentially identifying the leaf nodes and removing them one by one. The Jacobian of the score can be approximated by Stein gradient estimator with ridge RBF kernel regression \citep{rolland2022score}.

Once a topological order is estimated, the causal graph becomes constrained to be a sub-graph of a certain fully connected graph. However, it is necessary to prune this fully connected graph to remove spurious edges. This study uses the CAM pruning process to complete the step.

\subsection{CAM Pruning}

The above approaches control confounding variables by retaining key confounders during variable selection, removing irrelevant variables through correlation analysis. This section refining the causal graph via CAM pruning to eliminate spurious relationships while preserving causal integrity.

After arranging the nodes, the graph is refined by using the CAM pruning process, which removes unnecessary connections to reveal the actual causal structure, aligning with methods discussed by Rolland et al. 2022\citep{rolland2022score}. Detailed outputs are provided in the Appendix and include the following metrics \footnote{It is noted that "Variable Selection" is to ensure that important confounders are included before pruning begins, as removing key variables early can lead to residual confounding or spurious relationships. The formal analysis of correlation removes unrelated variables—those that have no meaningful relationship with the target variable or the other variables in the system. These variables are unlikely to act as confounders since they do not introduce residual confounding or spurious relationships when removed.

For validation, after CAM pruning, the causal structure is validated using domain expertise to ensure the robustness of the inferred causal graph. CAM pruning is not a substitute for confounding control methods. It is suggested to be used in combination with other techniques to ensure the validity of causal inferences. This is also the rationale for incorporating LLMs with expertise knowledge for further exploration.}:

\begin{itemize}
    \item Structural similarity: Evaluated using SID and SHD.
    \item Predictive accuracy: Measured through precision, recall, and F1 score.
\item Overall deviation: Assessed using L2 distance.
\end{itemize}

The graph (See Appendix Figure \ref{fig:FigA4}) highlights a structured approach to understanding how specific social factors influence carbon emissions and climate change. By focusing on the most influential variables—access to clean fuels in rural and urban areas and managing urban population growth—strategic decisions and policies can be more effectively targeted.

\section{Validation: From Correlations to Causation via LLM Inquiries}

In the specific context of climate change, do LLMs offer better causal inference? To address the request involving the exploration of causality factors for carbon emissions using the World Bank variables ("EG.CFT.ACCS.RU.ZS", "EG.CFT.ACCS.UR.ZS", "SP.URB.TOTL.IN.ZS") as the piror benchmark, the study categorizes questions into five main types for LLMs prompts (See Taxonomy of Causality). 

This study follows \citep{ceraolo2024}'s CausalQuest database focusing more on the economy and climate change background. The study follows \citep{ceraolo2024}'s CausalQuest database, but focuses more on the economic and climate change context. Similarly, the study adopts Pearl’s Causal Hierarchy (PCH) framework (\citep{pearl2018why,bareinboim2022pearl}), and defines a causal question as one that meets the following criteria: a question is considered causal if it involves, or if its solution process includes, any inquiry into the effects given a specific cause, and the causes given a specific effect, or the causal relationship between the given variables.
\subsection{Taxonomy of Causality}
The causal taxonomy-"Direct, Preventative, Facilitative, Resultative, and Influential"-describes various types of causal relationships that verbs can imply. This approach controls for confounding variables during LLM inquiries by leveraging a structured causal taxonomy to identify, classify, and account for different types of causal relationships\citep{liang2023holisticevaluationlanguagemodels,cui2024odysseycommonsensecausalityfoundational}.

The taxonomy classifies causal relationships into "Direct, Preventative, Facilitative, Resultative, and Influential" categories, ensuring that the LLMs recognize the nature of relationships between variables. By explicitly categorizing verbs that describe causal interactions, it helps avoid misinterpretation of ambiguous or indirect relationships, which could otherwise lead to confounding. In the case of carbon emissions:
A variable like access to clean technology might be classified as Facilitative (like "facilitates a reduction in emissions") rather than Direct (like "directly reduces emissions"), ensuring proper distinction. Urbanization, classified as Influential (like "influences emissions through energy use patterns"), ensures its role as a broader contextual factor is not conflated with a direct cause.

"The direct" refers to actions or driven forces that have a straightforward and immediate impact on an outcome. In this condition, the cause directly influences the effect without intermediary steps. For example, "increase" or "trigger" are direct causal verbs because they indicate a direct cause-effect relationship\citep{girju2002mining,kozareva2012cause,riaz2014recognizing,nazaruka2020overview}. Example: "Urban access to clean fuels directly reduces carbon emissions."

"The preventative" involves actions or causes that prevent or reduce the likelihood of a particular outcome. These verbs imply that the cause acts as a barrier to a negative effect. Common verbs include "prevent", "reduce" and "inhibit"\citep{allen2005local}. Example: "Improved access to clean technologies prevents an increase in carbon emissions."

"The facilitative" includes causes that make it easier or more likely for an effect to occur but do not directly cause the effect themselves. Facilitative causes provide support or create conditions that enable the outcome. Verbs like "enable", "allow", or "support" are examples\citep{harvey2002getting,wolff2003direct}. Example: "Access to urban clean fuels facilitates a reduction in carbon emissions."

"The resultative" describes causes that lead to specific outcomes, often emphasizing the result or consequence of an action. These verbs focus on the outcome rather than the action itself. Verbs like "lead to", "result in" or "cause" fit into this category\citep{boas2000resultative,pena2015constructionist}. Example: "The urban population increase results in higher carbon emissions."

"The influential" includes actions or factors that exert an influence on the effect but might not completely determine it. These causes often affect the likelihood or intensity of the effect indirectly. Verbs like "influence", "impact" or "affect" are often used\citep{yee1996causal,slovic2007affect,slovic2013risk}. Example: "Urbanization influences carbon emissions through changes in energy use patterns."

\section{Results of Causal Relationship and Interpretations}

Based on data of carbon emissions and social impacts (266 countries/regions, 70 socio-economic indicators, 20 years), the analysis identifies that Access to Clean Fuels and Technologies for Cooking (percent of Rural Population)("EG.CFT.ACCS.RU.ZS"), Access to Clean Fuels and Technologies for Cooking (percent of Urban Population)("EG.CFT.ACCS.UR.ZS"), and Urban Population (percent of Total Population)("SP.URB.TOTL.IN.ZS") have strong causal effects on carbon emissions per capita. These findings are consistent with previous research, which has verified the significant influence of these variables in the context of climate change and energy transitions.

Access to clean fuels and technologies for cooking, rural("EG.CFT.ACCS.RU.ZS"): This variable is highly influential in the context of climate change as it directly affects carbon emissions through the use of clean versus polluting energy sources in rural areas. A higher score indicates that rural access to clean fuels significantly reduces carbon emissions, highlighting its critical role in mitigating climate change impacts in less urbanized social context \citep{nathaniel2021environmental, verma2021energy}.

 Access to clean fuels and technologies for cooking, urban("EG.CFT.ACCS.UR.ZS"): Similar to the rural access variable, this factor measures the availability of clean cooking technologies in urban areas. Urban access is crucial since densely populated regions can contribute substantially to emissions. Improving clean energy access in urban areas can lead to a significant reduction in overall emissions, making it a key target for policy interventions \citep{naeem2023transitioning}.

 Urban population as a percentage of total population("SP.URB.TOTL.IN.ZS"): This variable captures the influence of urbanization on carbon emissions. As urban populations grow, the demand for energy, transportation, and industrial activity increases, contributing to higher emissions\citep{hankey2010impacts, madlener2011impacts,li2015impacts}. The score associated with this variable indicates that urbanization plays a major role in driving climate change, necessitating targeted strategies to manage urban growth sustainably.
\section {Conclusion: Evaluations and Integrations}

The three-step causal inference framework for data-driven decision-making in climate change context integrates correlation analysis, machine Learning, and LLM-interpretations. In this framework, correlation analysis helps narrow down and identify connections, causality provides a stricter and more precise understanding of these relationships, and LLMs interpret the results within specific scenarios.

Correlation provides a preliminary view of the relationships by highlighting mutual associations among variables and measures that indicate the extent to which two or more variables paired with each other. It narrows the scope of investigation by identifying potential connections between variables, but they do not provide insights into the nature of these connections.

Causality involves understanding the directional influence from one variable to another. In this research, causality goes a step beyond correlation by aiming to establish a cause-and-effect relationship between variables. Exploring and understanding causality is more stringent and complex because it requires not just observing that two variables occur together but also demonstrating that one variable produces an effect on another. 

LLMs could not handle causality explicitly and could not differentiate between mere correlations and true causal relationships (\citep{zečević2023causalparrotslargelanguage,liu2024large}. However, in interpreting results, LLMs can offer insights that are conditioned on their training data and the scenarios they are designed to understand. This means that while LLMs can be adept at identifying patterns and generating responses based on correlations, their ability to correctly interpret causal relationships could be improved by leveraging a structured causal taxonomy.

\setlength{\bibsep}{1.5pt}  

\bibliography{custom.bib}

\appendix

\section{Three-step Framework for Causal Analysis}

This research proposes a three-step framework for causal inference that progresses from understanding correlations to establishing causality, and finally to interpreting these relationships via LLMs. This approach leverages the different methods and exploits their distinctive advances to align with the understanding of climate change issues. This structured approach helps to systematically explore and analyze causality factors associated with carbon emissions and translates data patterns into LLM-inquiry-driven interpretations, which aids in gaining deeper insights and more interpretable policymaking.

\begin{figure}[H]
    \centering
    \includegraphics[width=1\linewidth]{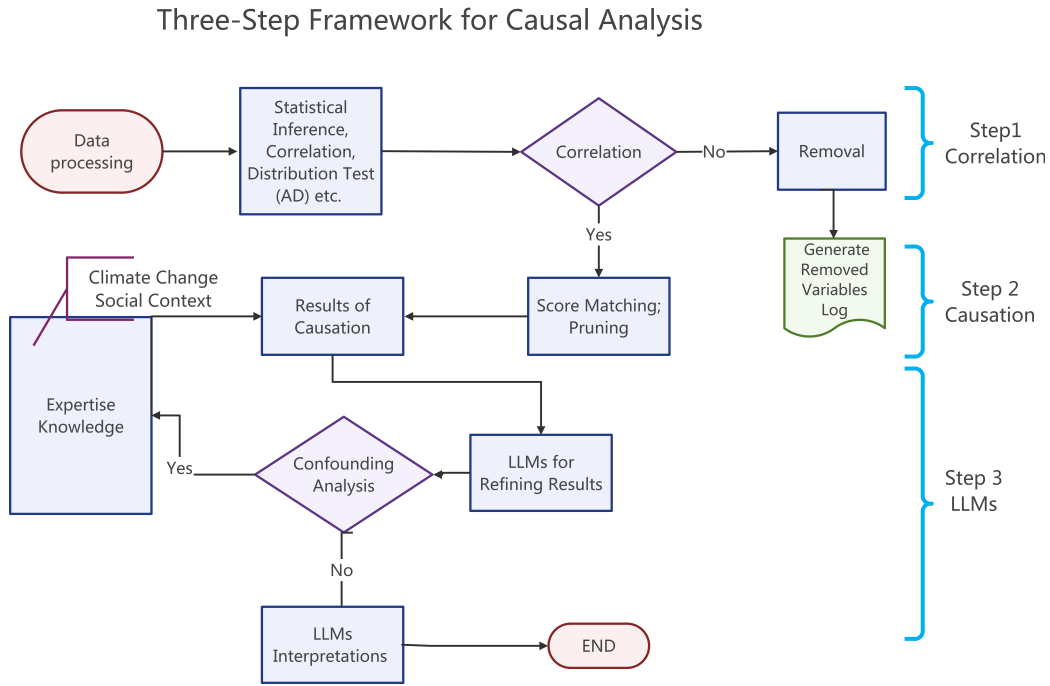}
    \caption{Three-step Framework for Causal Analysis}
    \label{fig:FigA1}
\end{figure}

\section{Setup and Main Results}
\begin{itemize}
    \item Data Availability
The data and code that support the findings of this study are available at https://github.com/shanshanfy/climate-change 

\item Developer Environment Availability
The author’s environment ‘ClimateChangePackages.yaml’ file for the Conda open-source package management system is provided through https://github.com/shanshanfy/climate-change. It allows isolated environments to manage packages without interference. The file contains the configuration of the project’s Python environment, including channels, dependencies, and library versions.
\end{itemize}

\subsection{Data Processing}

The research identifies the socioeconomic factors that influence and contribute to carbon emissions and climate change. The data is available at
\url{https://www.climatewatchdata.org/ghg-emissions}.
Total carbon emissions are measured as carbon emissions per capita. The complete carbon emission dataset is collected from 265 countries and includes 100 variables related to carbon emissions for the years 2000, 2005, 2010, 2015, and 2020. Emissions data are sourced from Climate Watch Historical GHG Emissions (1990-2020). 2023. Washington, DC: World Resources Institute.

\newpage
\section{Correlation Map}
\begin{figure}[h]
    \centering
    \includegraphics[width=1\linewidth]{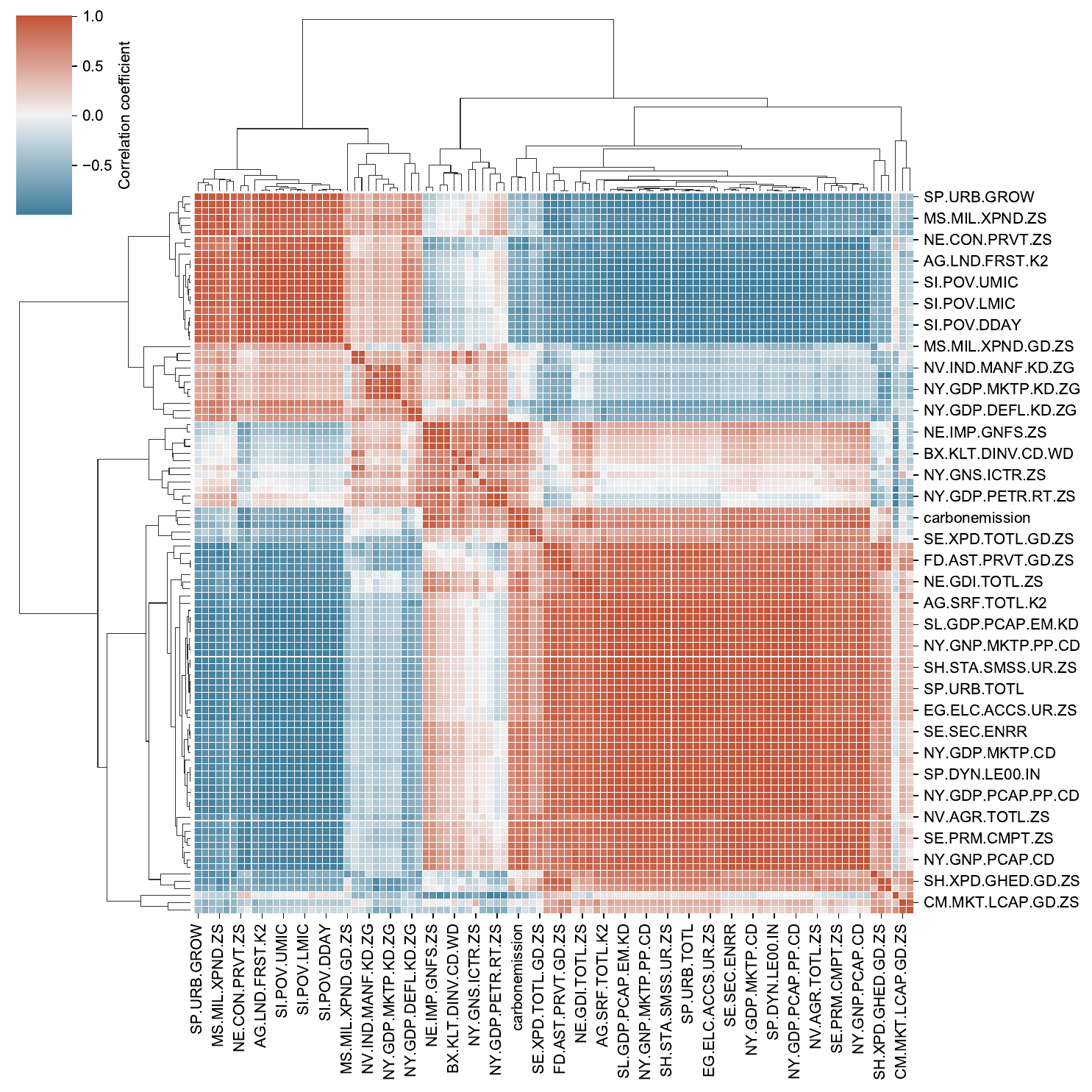}
    \caption{\textbf{Clustered Correlation Heatmap of Social Factors Influencing Carbon Emissions} This heatmap illustrates the correlations between various social factors and carbon emissions, highlighting key relationships. The clustering visually groups factors with similar correlation patterns, aiding in identifying which socioeconomic indicators most strongly influence carbon emissions, thereby providing insights into the complex interplay between social behavior and climate impact. The dendrogram, shown as lines on the top and left of the heatmap, represents hierarchical clustering. It groups variables based on the similarity of correlation or distance, with shorter line heights indicating higher similarity. Variables in the rows and columns are grouped to identify clusters with closely related pairwise relationships.}
    \label{fig:figA2}
\end{figure}

\subsection{Ordered Feature Correlation with Carbon Emissions}

\newpage

\section{Causal Analysis}

\subsection{Data Descriptions}

Given the substantial amount of missing data, the experiment eliminated columns with a missing ratio greater than 40\%, as well as some columns with data that are difficult to observe. The results have identified 15 studied variables related to the carbon emission variable, including "EG.CFT.ACCS.ZS", "EG.CFT.ACCS.RU.ZS", "EG.CFT.ACCS.UR.ZS", "EG.ELC.ACCS.UR.ZS", "EG.ELC.ACCS.ZS", "SP.URB.TOTL", "SP.URB.TOTL.IN.ZS", "SP.URB.GROW", "SE.SEC.DURS", "EG.FEC.RNEW.ZS", "SP.RUR.TOTL.ZS", "SP.RUR.TOTL", "AG.LND.FRST.ZS", "ER.FSH.CAPT.MT".The description of these variables is shown in Table 1.

Subsequently, the study treats the data in a numerical matrix \(V^t \in \mathbb{R}^{134 \times 15}\) and denotes EN.ATM.CO2E.PC by \(Y^t \in \mathbb{R}^{134 \times 1}\), where \(t \in \{2000, 2005, 2010, 2015, 2020\}\) denotes the year in which the data were collected. We then normalize and standardize each column of data. Finally, using a five-year interval as a step, the study investigates the causal relationship between \(X\) and \(Y\). The integrated observational data are as follows:

\[ D = \{ (V^{2000}, Y^{2005}), (V^{2005}, Y^{2010}), (V^{2010}, Y^{2015}), (V^{2015}, Y^{2020}) \}, \]

\begin{table}[h]
\centering
\caption{Description of the Studied Variables for Causal Analysis}
\label{tab:variables}
\begin{tabular}{|l|p{10cm}|}
\hline
\textbf{Variable} & \textbf{Description} \\ \hline
EG.CFT.ACCS.ZS & Access to clean fuels and technologies for cooking (\% of population) \\ \hline
EG.CFT.ACCS.RU.ZS & Access to clean fuels and technologies for cooking, rural (\% of rural population) \\ \hline
EG.CFT.ACCS.UR.ZS & Access to clean fuels and technologies for cooking, urban (\% of urban population) \\ \hline
EG.ELC.ACCS.UR.ZS & Access to electricity, urban (\% of urban population) \\ \hline
EG.ELC.ACCS.ZS & Access to electricity (\% of population) \\ \hline
SP.URB.TOTL & Urban population \\ \hline
SP.URB.TOTL.IN.ZS & Urban population (\% of total population) \\ \hline
SP.URB.GROW & Urban population growth (annual \%) \\ \hline
SE.SEC.DURS & Secondary education, duration (years) \\ \hline
EG.FEC.RNEW.ZS & Renewable energy consumption (\% of total final energy consumption) \\ \hline
SP.RUR.TOTL.ZS & Rural population (\% of total population) \\ \hline
SP.RUR.TOTL & Rural population \\ \hline
AG.LND.FRST.ZS & Forest area (\% of land area) \\ \hline
ER.FSH.CAPT.MT & Capture fisheries production (metric tons) \\ \hline
EN.ATM.CO2E.PC & CO2 emissions (metric tons per capita) \\ \hline
\end{tabular}
\end{table}

\subsection{Scoring Matching Output: Causal Relationships Among Social Factors and Carbon Emissions}

\begin{figure}[H]
    \centering
    \includegraphics[width=1\linewidth]{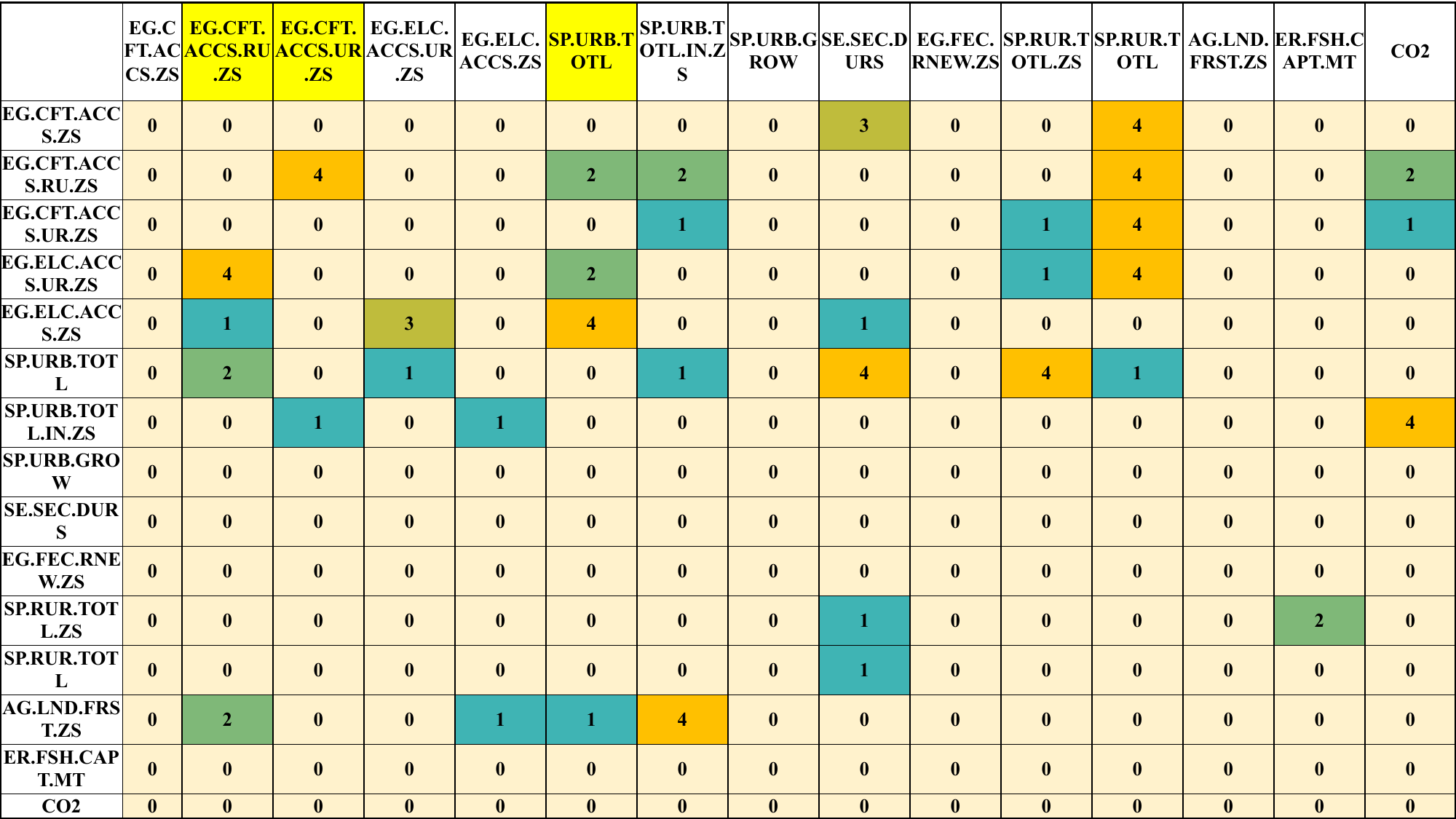}
    \caption{Causal Relationships Among Social Factors and Carbon Emissions. This scoring map illustrates the causal relationships between various social factors and carbon emissions, highlighting key variables: access to clean fuels and technologies for cooking in rural and urban areas (EG.CFT.ACCS.RU.ZS and EG.CFT.ACCS.UR.ZS) and urban population percentage (SP.URB.TOTL.IN.ZS). These factors show strong causal effects on carbon emissions per capita, emphasizing the interconnectedness of urbanization, energy use, and climate change.}
    \label{fig:FigA4}
\end{figure}

\subsection{CAM Pruning}

• Structural similarity (through SID and SHD).
• Predictive accuracy (using precision, recall, and F1 score).
•  Overall deviation (via L2 distance).


\subsubsection{Function Definition: BackRE}

The \texttt{backRE} function calculates several metrics to evaluate the accuracy of a predicted Directed Acyclic Graph (DAG) against the target DAG. Below is the Python implementation:

\begin{verbatim}
def backRE(tar_DAG, P_KCI):
    sid_val = SID(tar_DAG, P_KCI)
    shd_val = SHD(tar_DAG, P_KCI)
    precision, recall, f1 = f1_score(tar_DAG, P_KCI)
    distance = l2_distance(tar_DAG, P_KCI)
    return [sid_val, shd_val, precision, recall, f1, distance]
\end{verbatim}

The function computes:
\begin{itemize}
    \item \texttt{SID}: Structural Intervention Distance.
    \item \texttt{SHD}: Structural Hamming Distance.
    \item Precision, Recall, and F1-score to evaluate edge predictions.
    \item L2 Distance to measure overall deviation between the target DAG and the predicted DAG.
\end{itemize}

\subsubsection{L2 Distance Formula}
The L2 Distance is calculated using the formula:

\[
\text{L2 Distance} = \sqrt{\sum_{i,j} \left( A_{\text{true}}[i, j] - A_{\text{pred}}[i, j] \right)^2 }
\]

where:
\begin{itemize}
    \item \( A_{\text{true}}[i, j] \): Entry in the adjacency matrix of the true DAG.
    \item \( A_{\text{pred}}[i, j] \): Entry in the adjacency matrix of the predicted DAG.
\end{itemize}

This metric provides a scalar measure of the overall deviation between the true and predicted graphs.

\section{LLM Inquires}

\subsection{Related Works}
As public use of LLMs for tasks, various resources and tools have emerged to aid in prompt engineering and discovery\footnote{\url{https://platform.openai.com/docs/guides/prompt-engineering} \\ \url{https://huggingface.co/spaces/Gustavosta/MagicPrompt-Stable-Diffusion} \\ \url{https://promptomania.com/stable-diffusion-prompt-builder/}}. Instruction Categories provide different strategies for prompt engineering, and this study employs the following methods. Zero-shot Evaluation Instruction and Zero-shot-CoT Instruction are similar, with the latter explicitly incorporating "chain of thought" reasoning \citet{Brown2020, zhou2022large, Srivastava2022}. Both approaches assess the model's ability to apply its training to new, unseen tasks without prior specific examples. Few-shot Evaluation Instruction and Resample Instruction involve adaptive learning from a small set of examples or feedback, iteratively refining the prompts. Forward Generation predicts subsequent content based on the preceding context, commonly used in natural language generation. 

However, there is a lack of a comprehensive collection of causal questions in previous research \cite{ceraolo2024}. While related databases such as Google \citep{kwiatkowski2019natural}, Bing \citep{nguyen2016ms}, and questions posed to LLMs (H-to-LLM) from sources like ShareGPT and WildChat \citep{zhao2024wildchat} exist, none specifically focus on new sources of natural-causal questions, particularly causal questions directly asked to LLMs \citep{ouyang2022training,jin2024largelang}. Moreover, there is no dedicated database that addresses the context of climate change. 

\subsection{LLM-Generated Mixed Questions}

\textbf{\textit{Understanding Variables}}

\begin{itemize}
    \item \textbf{Direct:} What does \texttt{EG.CFT.ACCS.RU.ZS} represent in the context of global carbon emissions?
    \item \textbf{Influential:} How might urban access to clean fuels (\texttt{EG.CFT.ACCS.UR.ZS}) impact carbon emissions?
    \item \textbf{Facilitative:} What is the significance of \texttt{SP.URB.TOTL.IN.ZS} in studying urbanization effects on the environment?
    \item \textbf{Influential:} How do these variables interact to influence overall carbon emissions?
\end{itemize}

\textbf{\textit{Historical Data Analysis}}

\begin{itemize}
    \item \textbf{Resultative:} What trends are observable in \texttt{EG.CFT.ACCS.RU.ZS} over the last decade?
    \item \textbf{Resultative:} Has there been a significant change in \texttt{EG.CFT.ACCS.UR.ZS} data in major industrial countries?
    \item \textbf{Resultative:} How has the urban population percentage (\texttt{SP.URB.TOTL.IN.ZS}) changed in emerging economies?
    \item \textbf{Influential:} What historical events have significantly impacted these variables?
\end{itemize}

\textbf{\textit{Predictive Modeling}}

\begin{itemize}
    \item \textbf{Resultative:} Can we predict future trends in \texttt{EG.CFT.ACCS.RU.ZS} using past data?
    \item \textbf{Influential:} How might changes in \texttt{EG.CFT.ACCS.UR.ZS} predict shifts in urban carbon emissions?
    \item \textbf{Facilitative:} What models can forecast the growth of urban populations (\texttt{SP.URB.TOTL.IN.ZS})?
    \item \textbf{Preventative:} What are the potential future scenarios for these variables under different policy implementations?
\end{itemize}

\textbf{\textit{Policy Impact Evaluation}}

\begin{itemize}
    \item \textbf{Influential:} How have recent policies affected rural access to clean technologies (\texttt{EG.CFT.ACCS.RU.ZS})?
    \item \textbf{Resultative:} What are the environmental impacts of improved urban access to clean fuels (\texttt{EG.CFT.ACCS.UR.ZS})?
    \item \textbf{Influential:} How does urbanization measured by \texttt{SP.URB.TOTL.IN.ZS} correlate with national carbon emission policies?
    \item \textbf{Preventative:} What policy measures could potentially alter the trends in these variables most effectively?
\end{itemize}

\subsection{LLM-Generated Why Questions}

\textbf{\textit{Understanding Variables}}

\begin{itemize}
    \item \textbf{Direct:} Why does \texttt{EG.CFT.ACCS.RU.ZS} matter in the context of global carbon emissions?
    \item \textbf{Influential:} Why might urban access to clean fuels (\texttt{EG.CFT.ACCS.UR.ZS}) influence carbon emissions?
    \item \textbf{Facilitative:} Why is \texttt{SP.URB.TOTL.IN.ZS} significant when studying the effects of urbanization on the environment?
    \item \textbf{Influential:} Why do these variables interact to influence overall carbon emissions?
\end{itemize}

\textbf{\textit{Historical Data Analysis}}
\begin{itemize}
    \item \textbf{Resultative:} Why are there observable trends in \texttt{EG.CFT.ACCS.RU.ZS} over the last decade?
    \item \textbf{Resultative:} Why has there been a significant change in \texttt{EG.CFT.ACCS.UR.ZS} data in major industrial countries?
    \item \textbf{Resultative:} Why has the urban population percentage (\texttt{SP.URB.TOTL.IN.ZS}) changed in emerging economies?
    \item \textbf{Influential:} Why have certain historical events significantly impacted these variables?
\end{itemize}

\textbf{\textit{Predictive Modeling}}

\begin{itemize}
    \item \textbf{Resultative:} Why can past data on \texttt{EG.CFT.ACCS.RU.ZS} be used to predict future trends?
    \item \textbf{Influential:} Why might changes in \texttt{EG.CFT.ACCS.UR.ZS} predict shifts in urban carbon emissions?
    \item \textbf{Facilitative:} Why are certain models effective at forecasting the growth of urban populations (\texttt{SP.URB.TOTL.IN.ZS})?
    \item \textbf{Preventative:} Why could potential future scenarios for these variables differ under various policy implementations?
\end{itemize}

\textbf{\textit{Policy Impact Evaluation}}
\begin{itemize}
    \item \textbf{Influential:} Why have recent policies affected rural access to clean technologies (\texttt{EG.CFT.ACCS.RU.ZS})?
    \item \textbf{Resultative:} Why do improved urban access to clean fuels (\texttt{EG.CFT.ACCS.UR.ZS}) have environmental impacts?
    \item \textbf{Influential:} Why does urbanization, as measured by \texttt{SP.URB.TOTL.IN.ZS}, correlate with national carbon emission policies?
    \item \textbf{Preventative:} Why might certain policy measures most effectively alter the trends in these variables?
\end{itemize}

\section{Limitations and Discussions}

The study knocks on the door of causal and inference and evaluates the LLM-inquiry performance. However, understanding how to question causality within LLMs also involves recognizing the social norms embedded in human-machine interactions, as well as the social and moral dynamics present in language \citep{van2015detection,wang-etal-2018-modeling,forbes-etal-2020-social,cui2024odysseycommonsensecausalityfoundational}. These aspects are far more complex than simple data patterns.

Technically, three limitations exist. First, for data dependency, the accuracy and reliability of the causal inferences drawn from this framework depend heavily on the quality and completeness of the underlying data. Poor data quality or gaps can lead to incorrect conclusions, potentially misguiding important policy decisions.

Second, for model assumptions, the three-step framework relies on the global carbon emissions and climate change assumptions that may need to be supported across different scenarios or contexts, particularly in the complex, multifactorial climate change domain. Third, for generalizability, findings derived from this framework are context-specific and may not apply to different settings or scenarios without adjustments.

In future research, with more data and a deeper grounding in real-world societal settings, studies on vertical domains could be expanded on a larger scale and have a more profound impact on policymaking. 

\section{Acknowledgment}

The author thanks Anpeng Wu for his contributions to the analytical work on causal methodology and analysis. The author thanks Prof. Fei Wu for his comments and suggestions on causal analysis and Dr. Xin Qiu for his comments on the article. The author thanks the NeurIPS reviewers and the attendees of the Causality and Large Models workshop for their valuable comments.

\section{Ethical Considerations}
The research does not use any privacy-sensitive data. It uses a publicly available World Bank dataset that contains no information on ethical conflicts.
\end{document}